%% file: root.tex
\documentclass[letterpaper, 10 pt, conference]{ieeeconf}  

\IEEEoverridecommandlockouts                              

\usepackage{amsmath} 
\usepackage{amssymb}  
\usepackage{booktabs}
\usepackage[noadjust]{cite}
\usepackage[colorinlistoftodos,prependcaption]{todonotes}
\usepackage[caption=false,font=footnotesize]{subfig}
\expandafter\def\csname ver@transparent.sty\endcsname{}
\usepackage{svg}
\usepackage{float}
\usepackage[export]{adjustbox}
\usepackage{tikz}
\usepackage{tkz-euclide}
\usepackage{makecell}
\usepackage{pifont}

\graphicspath{{plots/}, {graphics/}, {figures/}, {logos/}, {NNs/PlotNeuralNet/pyexamples/}}
\usetikzlibrary{shadows.blur,calc,matrix,chains,positioning,decorations.pathreplacing,arrows,shapes.geometric}

\newcommand{\comment}[1]{}

\newcommand{\cmark}{\ding{51}}%
\newcommand{\xmark}{\ding{55}}%

\makeatletter 
\tikzstyle{inlinenotestyle} = [
    draw=\@todonotes@currentbordercolor,
    fill=\@todonotes@currentbackgroundcolor,
    line width=0.5pt,
    inner sep = 0.8 ex,
    rounded corners=4pt]

\renewcommand{\@todonotes@drawInlineNote}{%
        {\begin{tikzpicture}[remember picture,baseline=(current bounding box.base)]%
            \draw node[inlinenotestyle,font=\@todonotes@sizecommand, anchor=base,baseline]{%
              \if@todonotes@authorgiven%
                {\noindent \@todonotes@sizecommand \@todonotes@author:\,\@todonotes@text}%
              \else%
                {\noindent \@todonotes@sizecommand \@todonotes@text}%
              \fi};%
           \end{tikzpicture}}}%
\newcommand{\mytodo}[1]{\@todo[inline]{#1}}%
\makeatother

\newcommand{\degree}{$^{\circ}$}
\newcommand{\uvec}[1]{\hat{#1}}

\newcommand{\norm}[1]{\left\lVert#1\right\rVert_{2}}
\newcommand{\abs}[1]{\lvert#1\rvert}

\newcommand{\parenth}[1]{\left(#1\right)}

\newcommand{\PAR}[1]{{\bf #1:}}

\DeclareMathOperator*{\argmax}{arg\,max}

\newcommand{\featuremap}{F}
\newcommand{\upscalefac}{u}
\newcommand{\corrlayer}{C}
\newcommand{\corrdimI}{y}
\newcommand{\corrdimII}{x}
\newcommand{\corrdimIII}{w}
\newcommand{\transl}{t}
\newcommand{\quat}{q}
\newcommand{\rot}{R}
\newcommand{\scale}{s}

\newcommand{\corrindx}{l}
\newcommand{\corrprevindx}{p}
\newcommand{\poseloss}{\mathcal{L}_{pose}}
\newcommand{\auxloss}{\mathcal{L}_{aux}}

\newcommand{\hypo}{\mathcal{H}}
\newcommand{\quatlossweight}{\mu}
\newcommand{\scalelossweight}{\beta}
\newcommand{\camcenter}{c}
\newcommand{\image}{\mathcal{I}}
\newcommand{\unc}{\sigma}
\newcommand{\transform}{T}
\newcommand{\query}{q}
\newcommand{\refI}{i}
\newcommand{\refII}{j}
\newcommand{\refIII}{k}

\newcommand{\spatsize}{s}
\newcommand{\border}{b}
\newcommand{\matching}{M}
\newcommand{\ray}{v}
\newcommand{\raypos}{\lambda}

\newcommand{\intersection}{o}
\newcommand{\triangangle}{\alpha}
\newcommand{\scenecoordneighbor}{p_{max}}
\newcommand{\intersecmin}{\intersection_{max}}
\newcommand{\anglemax}{\gamma_{max}}
\newcommand{\distancemax}{d_{max}}
\newcommand{\auxpartloss}{e}

\makeatletter
\def\endthebibliography{%
  \def\@noitemerr{\@latex@warning{Empty `thebibliography' environment}}%
  \endlist
}
\makeatother

\title{\LARGE \bf
 Learning to Localize in New Environments from \\ Synthetic Training Data
}

\author{Dominik Winkelbauer \and Maximilian Denninger \and Rudolph Triebel
\thanks{German Aerospace Center (DLR), Institute of Robotics and Mechatronics, Dep. of Perception and Cognition, M\"unchner Str. 20, 82234 Wessling, Germany {\{\tt\small firstname.lastname\} at dlr.de}}
}

\begin{document}

\maketitle
\thispagestyle{empty}
\pagestyle{empty}

\begin{abstract}
\input{chapters/abstract}
\end{abstract}

\input{chapters/introduction}

\input{chapters/related_work}

\input{chapters/problem_description}

\input{chapters/pose_estimation_pipeline}

\input{chapters/experiments}

\input{chapters/conclusions}

\bibliographystyle{IEEEtran} 
\bibliography{bibliography}

\end{document}

%% file: chapters/abstract.tex
Most existing approaches for visual localization either need a
detailed 3D model of the environment or, in the case of learning-based
methods, must be retrained for each new scene. This can either be very
expensive or simply impossible for large, unknown environments, for
example in search-and-rescue scenarios. Although there are
learning-based approaches that operate scene-agnostically, the
generalization capability of these methods is still outperformed by
classical approaches.  In this paper, we present an approach that can
generalize to new scenes by applying specific changes to the model
architecture, including an extended regression part, the use of
hierarchical correlation layers, and the exploitation of scale and
uncertainty information.  Our approach outperforms the 5-point
algorithm using SIFT features on equally big images and additionally
surpasses all previous learning-based approaches that were trained on
different data. It is also superior to most of the approaches that
were specifically trained on the respective scenes.  We also evaluate
our approach in a scenario with only very few reference images,
showing that under such more realistic conditions our learning-based
approach considerably exceeds both existing learning-based and
classical methods.

%% file: chapters/introduction.tex
\section{INTRODUCTION}
\label{sec:introduction}

\begin{figure}
\vspace{-6mm}
\begin{center}
\subfloat[Query image (red)]{
  \begin{minipage}[b]{0.2\linewidth}
    \includegraphics[width=\linewidth,valign=c]{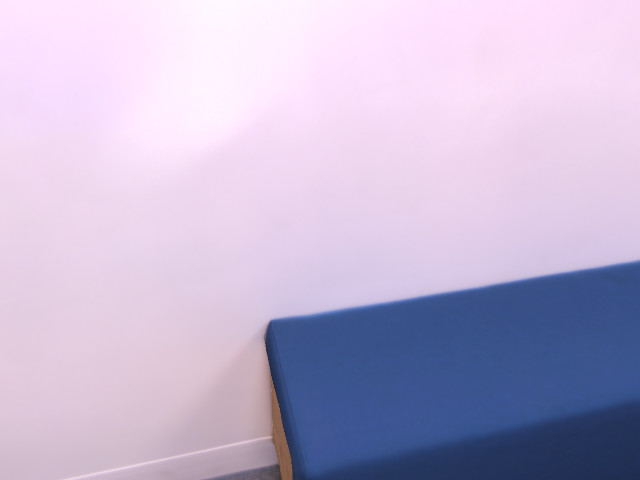}
  \end{minipage}
  }
  \subfloat[Scene]{
\begin{minipage}{0.51\linewidth}
\centering
\begin{minipage}{\linewidth}
\begin{center}
\input{figures/replica_examples/ex_1/latex_scene.tex}
\vspace{-0.45cm}
\end{center}
\end{minipage}
\end{minipage}
}
\newline
\centering
\vspace{0.2cm}
  \subfloat[Reference images (blue)]{
  \begin{minipage}[b]{\linewidth}
    \includegraphics[width=0.2\linewidth]{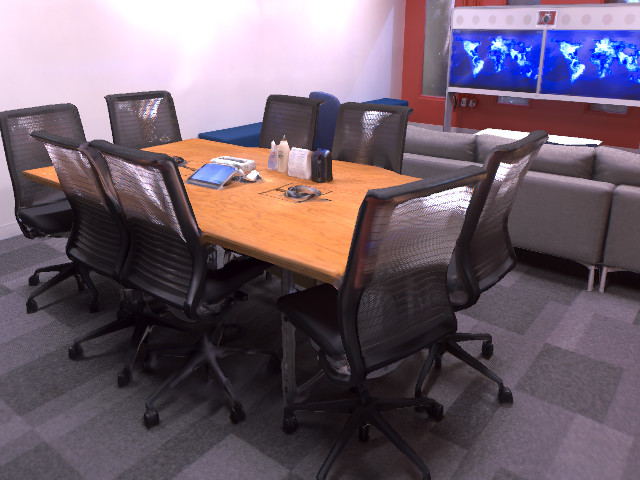}
    \includegraphics[width=0.2\linewidth]{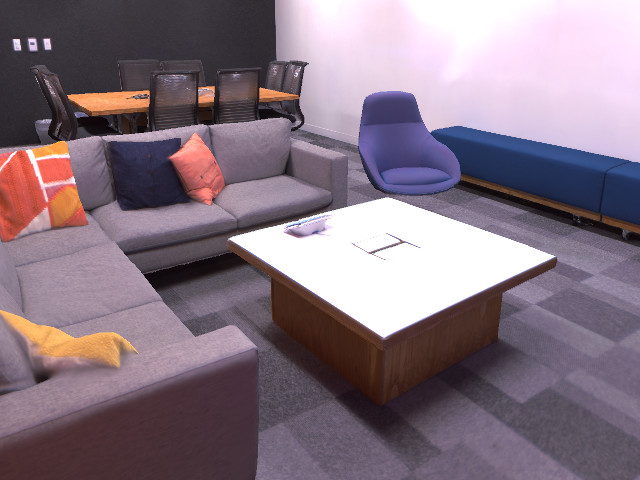}
    \includegraphics[width=0.2\linewidth]{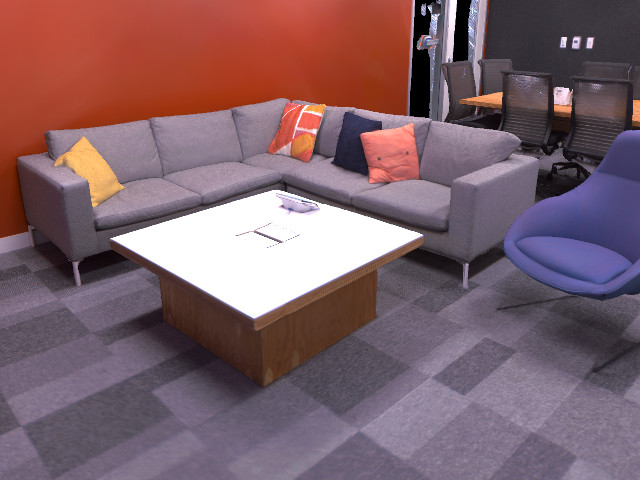}
    \includegraphics[width=0.2\linewidth]{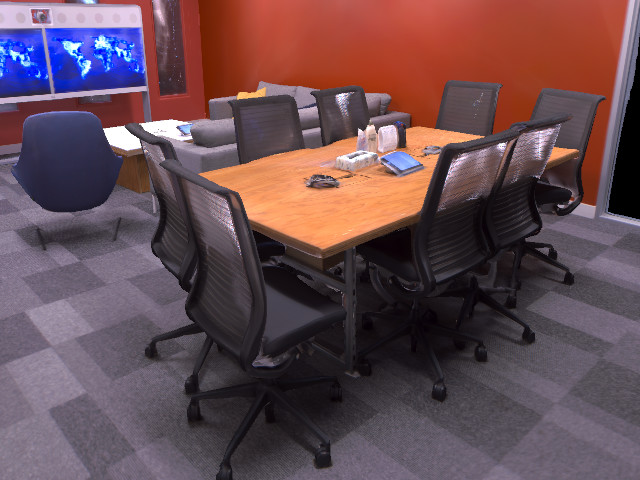}

  \end{minipage}
  }
\end{center}
\caption{Localization with reference images from only four given poses
  (blue triangles in (b), reference views in (c)). Note that finding
  an accurate pose from the query view in (a) is very difficult, but
  our approach makes a very good prediction (green triangle in (b)) to the ground truth pose (red).}
\label{fig:repl_example_hard}
\end{figure}

Robot localization is one of the major, fundamental tasks in robotics,
and it has been investigated already for several decades. Concretely,
the robot localization task consists of determining the coordinates of
the robot's position and orientation with respect to a global map
frame, based on its observed sensor input. To achieve this, some kind
of map representation must be available beforehand, which
distinguishes the localization task from the more general problem of
Simultaneous Localization and Mapping (SLAM). This has the major
advantage that the localization process itself is simpler,
because it only requires finding a ``match'' between the current
observation and the map representation. The downside however is that
building an accurate and useful map of the environment is itself an
nontrivial process, which is most often performed by again employing
SLAM techniques, albeit this is done offline, i.e. before actually
deploying the robot.
An additional problem of handling localization separately from mapping
is that, in terms of sensing modality the map representation must be
comparable to the observations made during localization, because
otherwise a good matching is difficult to achieve. For most
current robot systems, this means that mapping is done with the same
sensor setup and the same parameter set as is used during the actual
localization process. This implies that a robot that is to be deployed
in a new environment needs to build a map of that environment first
with its own sensor stack, although some map representation might
already be available.
In this paper, we are going to address these issues with a novel
approach. The only sensing modality we consider are standard RGB
cameras, which makes our framework very flexible and
versatile. Concretely, our map representation simply consists of a set
of reference images where each has an assigned 6D transform with
respect to the global map frame. Thus, to determine the robot's pose
we compare the currently observed image frame with all
reference images and estimate the transforms between the current pose
and the reference images using a deep learning architecture. 
Although there already exist similar approaches \cite{Laskar2017CameraRB,Balntas2018RelocNetCM,Zhou2019ToLO} , they still are inaccurate on scenes not contained in their training data.
We show that training our specific network
architecture only with synthetic data is
sufficient to localize a robot in an unknown, real environment, and
the accuracy of this learning-based localization outperforms classical model-based approaches.
Furthermore, we show that in the case where only a few reference images are available (see fig.~\ref{fig:repl_example_hard}), our method considerably outperforms classical methods.

%% file: figures/replica_examples/ex_1/latex_scene.tex
\begin{tikzpicture}
    \node[anchor=south west,inner sep=0] (image) at (0,0) {\includegraphics[width=0.9\textwidth]{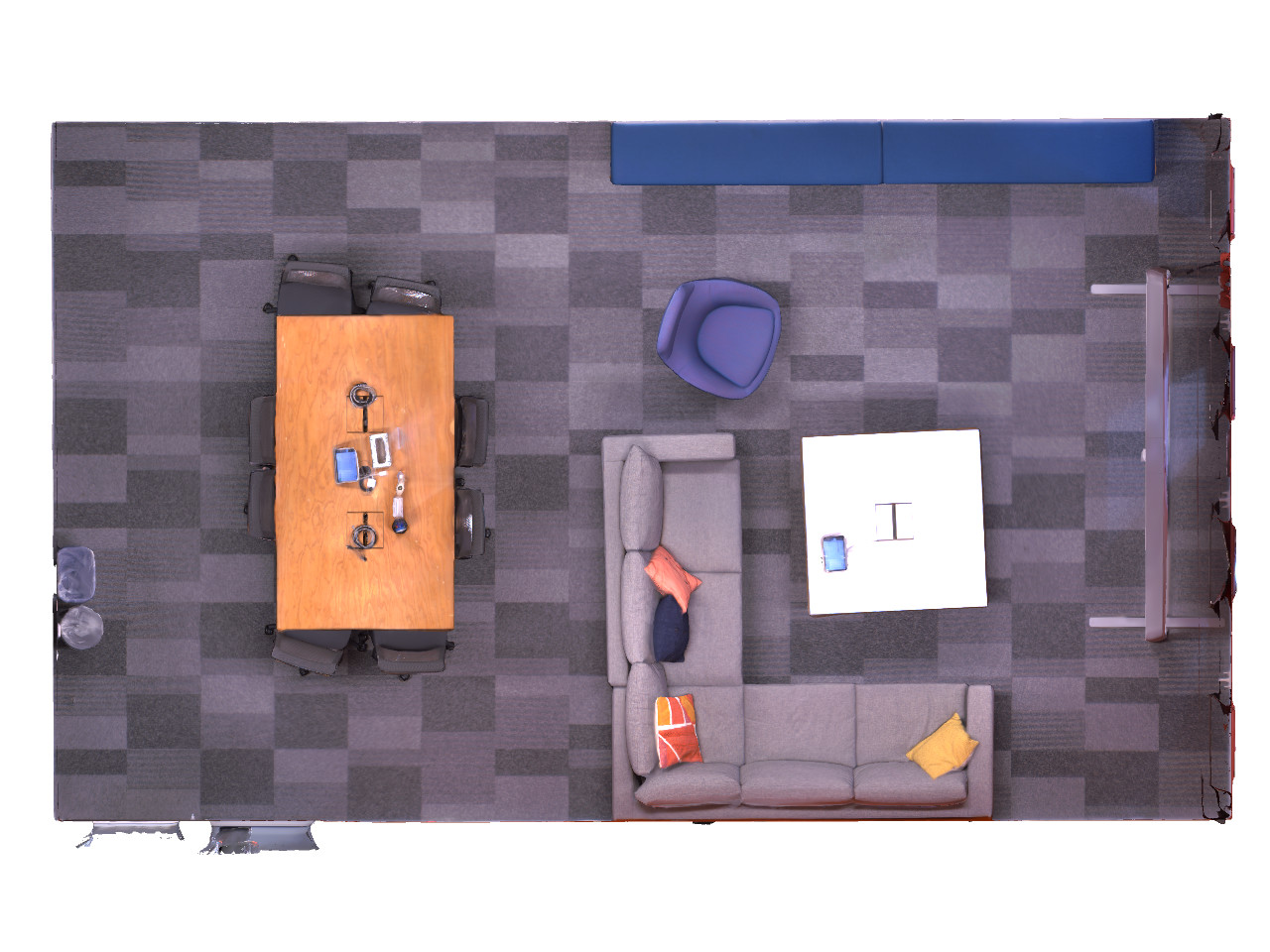}};
    \begin{scope}[x={(image.south east)},y={(image.north west)}]
        \fill[red](0.46703498731502324, 0.705724618347318) ellipse (0.01 and 0.013333);

\node at (0.46703498731502324, 0.705724618347318) {\includegraphics[scale=0.027, angle=280]{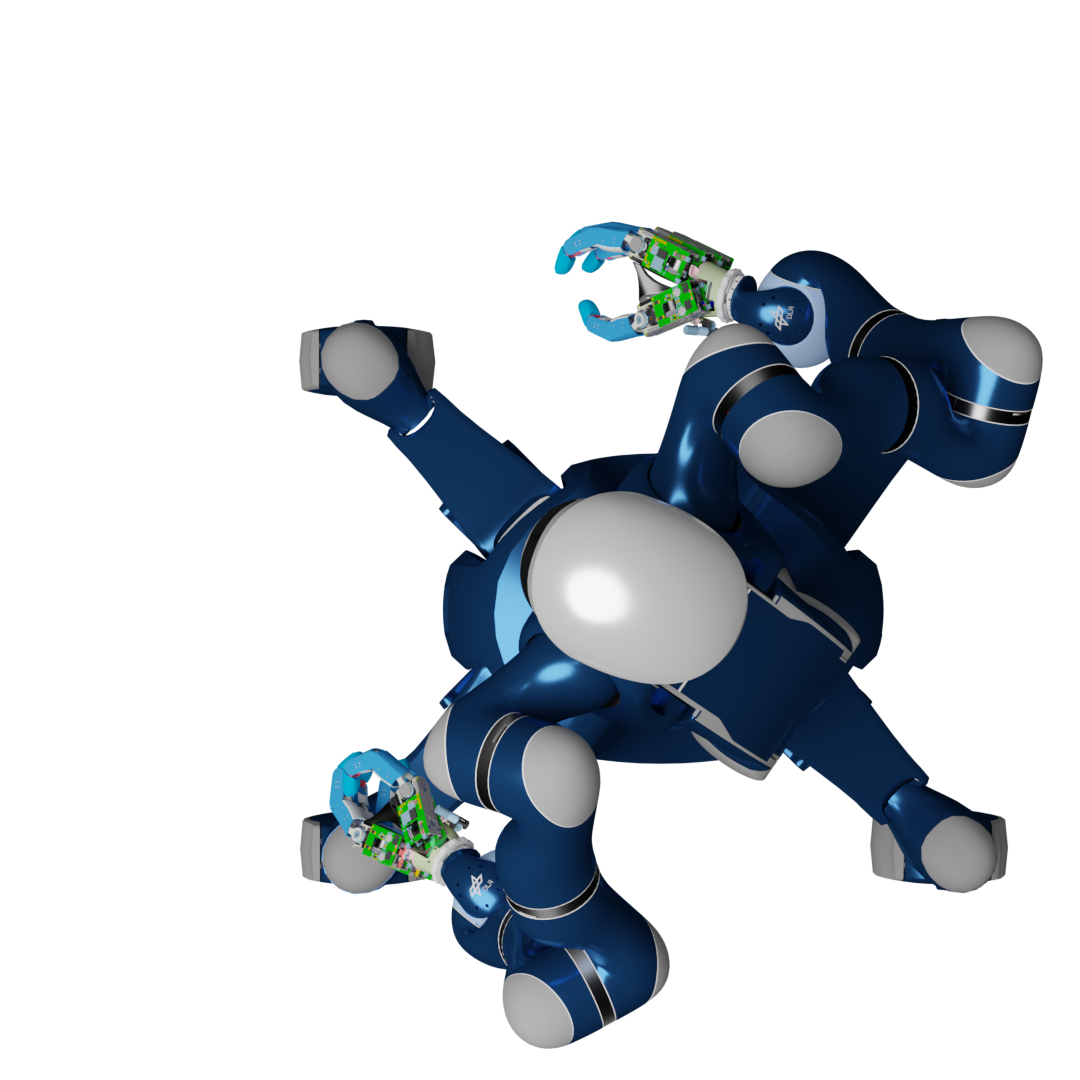}};
\draw[ fill=white, draw=red, very thick, fill opacity=0.5](0.46703498731502324, 0.705724618347318) - - (0.5235982817894567, 0.7837650040131945) - - (0.44462791122933715, 0.8100584875864507) - - cycle;

\fill[green](0.41681610797988056, 0.5966918425585794) ellipse (0.01 and 0.013333);
\draw[draw=green, very thick, fill=white, fill opacity=0.5](0.41681610797988056, 0.5966918425585794) - - (0.47858970759237834, 0.667361314875818) - - (0.40180175407313173, 0.7033565675000671) - - cycle;
\fill[blue](0.07536799074359313, 0.19823692610525118) ellipse (0.01 and 0.013333);
\draw[ fill=white, draw=blue, very thick, fill opacity=0.5](0.07536799074359313, 0.19823692610525118) - - (0.15593760929827982, 0.21365599091155807) - - (0.10563783398837445, 0.2989803041126796) - - cycle;
\fill[blue](0.9269051916918558, 0.17914109581314075) ellipse (0.01 and 0.013333);
\draw[ fill=white, draw=blue, very thick, fill opacity=0.5](0.9269051916918558, 0.17914109581314075) - - (0.8924466668751443, 0.27746306526732334) - - (0.8458139354877057, 0.18851283072629094) - - cycle;
\fill[blue](0.8984520424930642, 0.7839421915183463) ellipse (0.01 and 0.013333);
\draw[ fill=white, draw=blue, very thick, fill opacity=0.5](0.8984520424930642, 0.7839421915183463) - - (0.8202565738652885, 0.7538113703256931) - - (0.8789248686535472, 0.6785844901069601) - - cycle;
\fill[blue](0.07105377902302501, 0.8509629263931278) ellipse (0.01 and 0.013333);
\draw[ fill=white, draw=blue, very thick, fill opacity=0.5](0.07105377902302501, 0.8509629263931278) - - (0.0951820153887542, 0.747313743533035) - - (0.1504400652313203, 0.8269992358038007) - - cycle;

    \end{scope}
\end{tikzpicture}
                    

%% file: chapters/related_work.tex
\section{Related work}
\label{sec:related_work}

\subsection{Image retrieval}
\label{sec:related_ir}

Image retrieval can be used to get a rough but fast estimate of the pose of a given image \cite{sattler2017large}.
By using techniques like DenseVLAD \cite{torii201524}, NetVLAD \cite{arandjelovic2016netvlad} or Disloc \cite{arandjelovic2014dislocation} a compact representation is calculated for each image, then the pose of the image with the closest descriptor is used.
As the accuracy of image retrieval is low, more sophisticated approaches \cite{zhang2006image,Zhou2019ToLO}, including ours, further refine the pose estimate and use image retrieval only as a first step to reduce the search space.

\subsection{3D structure-based localization}
\label{sec:related_ir}

3D structure-based methods do pose estimation based on matches between points in the image and points in a 3D model of the scene.
Although this leads to accurate predictions, building such a 3D model requires that enough reference images are available.
Our approach can also handle situations with a low number of reference images.
There also exist learning-based approaches using CNNs \cite{cavallari2017fly, brachmann2017dsac, brachmann2018learning, li2018scene} or random forests \cite{cavallari2017fly, meng2018exploiting, massiceti2017random, brachmann2017dsac, valentin2015exploiting} which predict for every pixel of a given image the corresponding matching 3D coordinate.
Altough these approaches are more memory efficient during inference, they need a 3D model for training or in the case of DSAC++ \cite{brachmann2018learning} need at least expensive retraining for each new scene.

\subsection{Absolute pose regression}
\label{sec:related_abs_pose_regr}

Absolute pose regression methods are learning-based approaches that try to regress directly the absolute pose of a given query image.
\cite{DBLP:journals/corr/KendallGC15} initially proposed this idea in the form of PoseNet, a CNN based on a pretrained GoogLeNet \cite{szegedy2015going}.
Follow-up approaches have interchanged the GoogLeNet with various other backbones \cite{simonyan2014very, naseer2017deep,Valada2018DeepAL, Saha2018ImprovedVR,Melekhov2017ImageBasedLU, Walch2016ImageBasedLU}. 
Independent of the exact architecture, \cite{sattler2019understanding} discovered that absolute pose estimation models poorly generalize to query images from the same scene that are further away from the training images.
By training on a big synthetic dataset our approach does not overfit on specific camera constellations.
Bayesian PoseNet \cite{Kendall2015ModellingUI} is trained via monte carlo dropout which we also use in our network to improve the triangulation step.
Other approaches \cite{Brahmbhatt2017GeometryAwareLO, Valada2018DeepAL,Li2019RelativeGS,DBLP:journals/corr/abs-1804-08366} suggest to additionally train the network on auxiliary losses like semantic segmentation or relative pose estimation.
As there are always losses which focus on scene specific properties, a retraining for each new scene is still necessary. We do not suffer from this as we train only once.

\subsection{Relative pose regression}
\label{sec:related_rel_pose_regr}

Some prior methods \cite{Laskar2017CameraRB, Balntas2018RelocNetCM, ding2019camnet, Zhou2019ToLO} estimate the relative pose between the query image and close-by reference images whose absolute poses are known. Then, the absolute pose of the query image is triangulated.
Although these approaches are in principle built to generalize, they do not perform better than models that were specifically trained on these scenes\cite{sattler2019understanding,Zhou2019ToLO}. In contrast, we show for the first time that it is possible and even beneficial to generalize from unrelated data without retraining.
Our approach is most similar to EssNet \cite{Zhou2019ToLO}, however with some major differences, as we use an extended regression part and hierarchical correlation layers.
Also, we represent the relative pose as a translation direction and a quaternion, which removes the need of scene-specific weighting, while no projection layer is necessary and the predicted pose does not contain ambiguities, which is the case when using essential matrices.
Additionally, we require a smaller image size which makes our approach require less computation time, which is limited on mobile robots.

%% file: chapters/problem_description.tex
\section{Problem description}

Visual relocalization can be defined as estimating the absolute pose in the form of a homogeneous transformation matrix $\transform_\query \in \mathbb{R}^{4 \times 4}$ of a given query image $\image_\query$.
Thereby, only a set of reference images $\image_\refI$ together with their absolute poses $\transform_\refI$ is given.
We call the rotational part of such a transformation matrix then $\rot_\refI \in \mathbb{R}^{3 \times 3}$ and the translational part $\transl_\refI \in \mathbb{R}^3$.
Instead of estimating the absolute pose directly, relative pose estimation methods like our approach estimate the relative pose $\transform_{\refI \rightarrow \query}$ between the query image $\image_\query$ and one or multiple reference images $\image_\refI$ and combine these with the known absolute poses of the reference images to get an estimate for the absolute pose of the query image.

%% file: chapters/pose_estimation_pipeline.tex
\section{Pose estimation pipeline}

For a given query image we first retrieve five reference images, then estimate the relative pose between each query-reference pair and finally triangulate the absolute pose of the given query image.

\input{chapters/pose_estimation_pipeline/retrieval}

\input{chapters/pose_estimation_pipeline/rel_pose_estimation}

\input{chapters/pose_estimation_pipeline/triangulation}

%% file: chapters/pose_estimation_pipeline/retrieval.tex
\subsection{Image retrieval}

Similar to Zhou et al. \cite{Zhou2019ToLO}, we collect the five
closest reference images according to the Euclidean distance of the
respective DenseVLAD descriptors, while skipping images whose distance to
the camera centers of the already selected images is not in range $[0.05\text{m}, 10\text{m}]$.

%% file: chapters/pose_estimation_pipeline/rel_pose_estimation.tex
\subsection{Relative pose estimation}
\label{sec:rel_pose_est}

\begin{figure*}
\begin{minipage}{1\textwidth}
\centering
\input{figures/network.tex}
\end{minipage}
\caption{Visualization of our relative pose estimation network ExReNet. Inside the ResNet stacks the number of blocks and the feature sizes are noted.}
\label{fig:network}
\end{figure*}
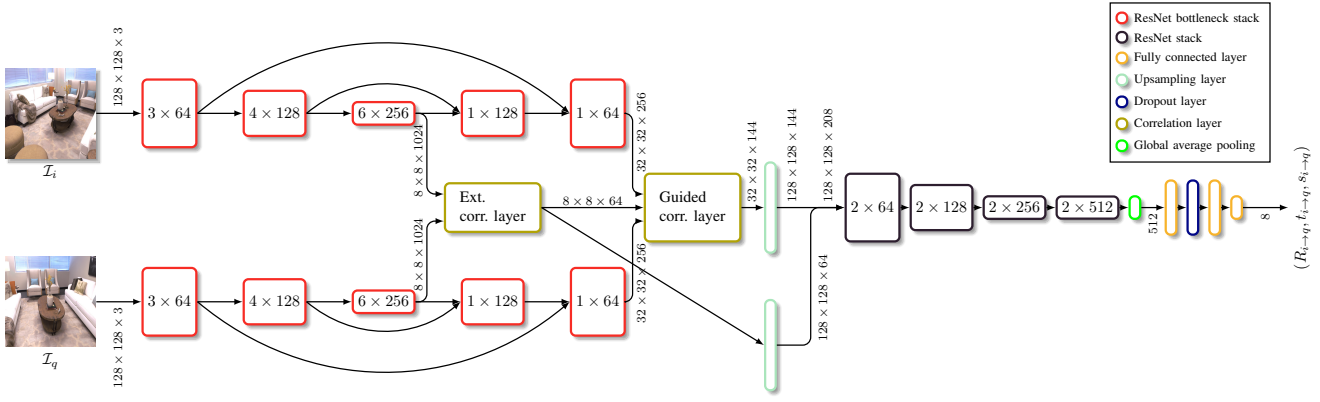

To estimate the relative pose between two given images we train a
siamese neural network, which we call ExReNet (see
Fig.~\ref{fig:network}). This consists of three stages: individual
feature extraction, feature matching, and relative pose regression.

\PAR{Feature extraction}
A ResNet50 pretrained on ImageNet is used up to its third stack to extract a feature map of $8 \times 8 \times 1024$ per image.
We further apply two upsampling steps in a UNet-like \cite{ronneberger2015u} fashion to obtain a second feature map of $32 \times 32 \times 256$ per image.

\PAR{Feature matching}
As in Zhou et al. \cite{Zhou2019ToLO} we use a correlation layer to combine both feature maps.
An extensive correlation layer $\corrindx$ computes for two given feature maps $\featuremap_1$ and $\featuremap_2$ with spatial resolution $s_\corrindx$ a three-dimensional correlation map $\corrlayer_\corrindx$ by performing the dot product between all possible feature vector combinations, i.e.
\begin{align}
\corrlayer_\corrindx(\corrdimI_1,\corrdimII_1,\corrdimIII) =&  \featuremap_1(\corrdimI_1,\corrdimII_1)^T \featuremap_2(\corrdimI_2,\corrdimII_2),
\end{align}
where $\corrdimIII = \corrdimII_2 \spatsize_\corrindx + \corrdimI_2$ and $\corrdimI_1,\corrdimII_1,\corrdimI_2,\corrdimII_2 \in \{0, ... ,\spatsize_\corrindx - 1\}$.

\begin{figure}
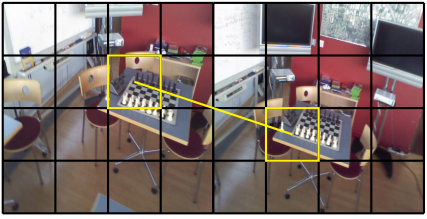
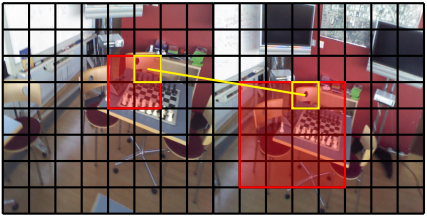

\subfloat[Extensive correlation layer on $4 \times 4$ feature maps]{
\begin{minipage}{1\linewidth}
\centering
\vspace{2mm}
\begin{tikzpicture}[scale=0.7, every node/.style={scale=0.7}]
\node[inner sep=0pt] () at (0,0)
    {\includegraphics[width=4cm,height=4cm]{q.jpg}};
    
  \foreach \x in {-2,...,2}
       {\draw[-,thick] (\x,2) -- (\x,-2);} 
  \foreach \y in {-2,...,2}
       {\draw[-,thick] (2,\y) -- (-2,\y);} 

\node[inner sep=0pt] () at (4,0)
    {\includegraphics[width=4cm,height=4cm]{r4.jpg}};

  \foreach \x in {-2,...,2}
       {\draw[-,thick] (\x+4,2) -- (\x+4,-2);} 
  \foreach \y in {-2,...,2}
       {\draw[-,thick] (6,\y) -- (2,\y);} 
       
   \draw[-,thick,draw=yellow] (0.5,0.5) -- (3.5,-0.5);
   
   \draw[draw=yellow, thick] (0,0) rectangle ++(1,1);
   
   \draw[draw=yellow, thick] (3,-1) rectangle ++(1,1);

\end{tikzpicture}
\end{minipage}
}
\newline
\subfloat[Correlation layer on $8 \times 8$ feature maps guided by the results from the correlation layer above using a border $b$ of $0.5$]{

\begin{minipage}{1\linewidth}
\centering
\begin{tikzpicture}[scale=0.7, every node/.style={scale=0.7}]
\node[inner sep=0pt] () at (0,0)
    {\includegraphics[width=4cm,height=4cm]{q.jpg}};
    
  \foreach \x in {-4,...,4}
       {\draw[-,thick] (\x/2,2) -- (\x/2,-2);} 
  \foreach \y in {-4,...,4}
       {\draw[-,thick] (2,\y/2) -- (-2,\y/2);} 

\node[inner sep=0pt] () at (4,0)
    {\includegraphics[width=4cm,height=4cm]{r4.jpg}};
    
  \foreach \x in {-4,...,4}
       {\draw[-,thick] (\x/2+4,2) -- (\x/2+4,-2);} 
  \foreach \y in {-4,...,4}
       {\draw[-,thick] (6,\y/2) -- (2,\y/2);} 
       
   \draw[fill=red, thick, fill opacity=0.35, draw=red] (0,0) rectangle ++(1,1);
   
   \draw[fill=red, thick, fill opacity=0.35, draw=red] (2.5,-1.5) rectangle ++(2,2);
       
   \draw[-,thick,draw=yellow] (0.75,0.75) -- (3.75,0.25);
   
   \draw[draw=yellow, thick] (0.5,0.5) rectangle ++(0.5,0.5);
   
   \draw[draw=yellow, thick] (3.5,0) rectangle ++(0.5,0.5);
   
\end{tikzpicture}
\end{minipage}
}
\caption{Visualizes two correlation layers whereby one is guided by the other: In each layer one correct matching is visualized in yellow. The guidance of the upper corr. layer is visualized in red in the lower one.}
\label{fig:guided_correlation}
\end{figure}

We found that, especially if the baseline between the two images is small, having higher resolution feature matches improves the relative pose estimates drastically.
However, for a given image side length $\spatsize_\corrindx$ the extensive correlation layer performs $\spatsize_\corrindx^4$ dot products, which makes it computationally infeasible for higher resolutions.
To counteract this issue we propose to apply multiple correlation layers in a hierarchical fashion:
Thereby, only the first low resolution correlation layer performs extensive matching, while the following correlation layers are guided by the matchings produced at the previous layer.
For a given feature vector a guided correlation layer only looks for potential matches in the region where the match of the corresponding coarse feature vector in the previous correlation layer has been found.
Fig.~\ref{fig:guided_correlation} visualizes this concept using an example image pair from ``7-Scenes''.
In detail, a correlation layer $\corrindx$ guided by the previous correlation layer $\corrprevindx$ is defined as follows:
\begin{equation}
\begin{split}
\corrlayer_\corrindx(\corrdimI_1,\corrdimII_1,\corrdimIII) = \featuremap_1(\corrdimI_1,\corrdimII_1)^T \featuremap_2(&(\matching_\corrdimI(\corrdimI_1,\corrdimII_1)- \border_\corrindx) \upscalefac_\corrindx + \corrdimI_2  , \\ &(\matching_\corrdimII(\corrdimI_1,\corrdimII_1) - \border_\corrindx) \upscalefac_\corrindx + \corrdimII_2  )
\end{split}
\end{equation}
where the size of the search area is $d_\corrindx = \upscalefac_\corrindx (1 + 2 \border_\corrindx) $ and the channel coordinate $\corrdimIII = \corrdimII_2 d_\corrindx + \corrdimI_2$.
Feature coordinates are defined as $\corrdimI_1, \corrdimII_1 \in \{0, ..., \spatsize_\corrindx - 1\}$ and $\corrdimI_2, \corrdimII_2 \in \{0, ..., d_\corrindx - 1\}$.
The border $b_l$ is added around the search area to increase the number of possible correct matches.
The upscale factor $\upscalefac_\corrindx = \spatsize_\corrindx / \spatsize_\corrprevindx$ describes the change in spatial resolution between the current correlation layer $\corrindx$ and the previous one $\corrprevindx$.
The guidance from the previous layer $p$ is represented as the corresponding match $\matching$:
\begin{align}
\matching_\corrdimI(\corrdimI,\corrdimII) \equiv&  \corrdimIII_{max}(\corrdimI,\corrdimII) \mod \spatsize_p \quad \mbox{or}\\
\matching_\corrdimII(\corrdimI,\corrdimII) =& \lfloor \frac{\corrdimIII_{max}(\corrdimI,\corrdimII) }{\spatsize_p} \rfloor
\end{align}
where $\corrdimIII_{max}(\corrdimI,\corrdimII)  = \argmax_\corrdimIII{\corrlayer_\corrprevindx(\lfloor \corrdimI / \upscalefac_\corrindx \rfloor,\lfloor \corrdimII / \upscalefac_\corrindx \rfloor, \corrdimIII)}$.

We found that training a network using hierarchical correlation layers with only the pose regression loss does not fully make use of the network's potential.
Even if the non-differentiable argmax is replaced by an equivalent softmax, the training gets stuck in a local minimum, such that the network only makes use of the high-level extensive correlation layer and neglects any further guided correlation layers.
To solve this issue, we train our network using an additional auxiliary loss, that makes sure the feature combinations with maximum dot products correspond to actual correct feature matches.
Using depth information we compute for every pixel in the first image the corresponding matching pixel in the second image.
Afterwards, we can compute for every feature vector combination $\featuremap_1(\corrdimI_1,\corrdimII_1), \featuremap_2(\corrdimI_2,\corrdimII_2)$ the number of pixels $n(\corrdimI_1,\corrdimII_1,\corrdimIII)$ that are included in the spatial region covered by $\featuremap_1(\corrdimI_1,\corrdimII_1)$ and whose matching pixel from the other image is covered by $\featuremap_2(\corrdimI_2,\corrdimII_2)$.

For an extensive correlation layer $\corrindx$ the auxiliary loss $\auxloss^\corrindx(\corrdimI,\corrdimII,\corrdimIII)$ for a given feature combination is defined as
\begin{equation}
\auxloss^\corrindx(\corrdimI,\corrdimII,\corrdimIII) =\frac{1}{\sum_q \delta_{\corrdimIII,q}} \sum_q \delta_{\corrdimIII,q} \cdot \auxpartloss(\corrdimI,\corrdimII,\corrdimIII,q) 
\end{equation}
\begin{equation}
\auxpartloss(\corrdimI,\corrdimII,\corrdimIII,q) = \max(0, \corrlayer_l(\corrdimI,\corrdimII,q) - \corrlayer_l(\corrdimI,\corrdimII,\corrdimIII) + 1) 
\end{equation}
\begin{equation}
\delta_{\corrdimIII,q} =  \begin{cases}
     1 & \text{if } n_l(\corrdimI,\corrdimII,\corrdimIII) > 0 \text{ and } n_l(\corrdimI,\corrdimII,q) = 0 \\
     0 & \text{otherwise}. 
    \end{cases} \\
\end{equation}
We formulate the auxiliary loss in the form of a triplet loss, where positive pairs are feature matches that correspond to correct pixel matches and negative pairs are matches that correspond to no actual matches.
For the guided correlation layer the auxiliary loss can be formulated accordingly.
The total auxiliary loss is defined as the mean over all correlation layers and across all their feature combinations.

In ExReNet we use one extensive corr. layer applied to $8 \times 8$ feature maps and one guided corr. layer applied to $32 \times 32$ using a border $b = 1$.
Compared to using only one corr. layer applied to $32 \times 32$, this reduces the total training memory consumption by   32\% and training time by 19\% while being more accurate as higher dimensional feature vectors are taken into account.

\captionsetup[subfigure]{justification=justified,singlelinecheck=false}

\captionsetup[subfigure]{justification=centering}

\PAR{Pose regression}
In ExReNet we first upsample the feature matches of both corr. layers to  $128 \times 128$, concatenate them along the channel dimension and then apply a ResNet18.
The spatial dimension is reduced afterwards via a global average pooling layer, followed by three fully-connected layers: Two with size 512 and one to map the features to the output vector which has a length of seven.
This output consists of a three dimensional vector representing the translation direction $\uvec{\transl}$ and a four dimensional quaternion $q$.

Our regressor has substantially more layers compared to existing architectures\cite{Kendall2017GeometricLF,Balntas2018RelocNetCM,Zhou2019ToLO}, as we found that especially layers that operate before collapsing the spatial dimension increase the generalization capability to new scenes.

\PAR{Loss function}
The used loss consists of one L1-loss for translation and one for rotation:
$\poseloss = \abs{\uvec{\transl} - \uvec{\transl^*}} + \quatlossweight \abs{\quat - \quat^*}$
In contrast to previous approaches, we train our network to predict the inverse relative pose, as the inverse is also used in the triangulation step:
So, for a given image pair $(\image_\refI, \image_\query)$ the ground truth relative translation direction $\uvec{\transl}^*$ equals $-\rot_{\refI \rightarrow \query}^T \uvec{\transl}_{\refI \rightarrow \query}$ and the ground truth relative rotation $\quat^*$ corresponds to $\rot_{\refI \rightarrow \query}^T$.
As we do not learn the fully-scaled translation, but only the normalized direction vector, no scene-specific weighting between translation and rotation loss is required, and as both parts have the same scale, we found that $\quatlossweight = 1$ performs well.

\PAR{Scale estimation}
Previous methods \cite{Laskar2017CameraRB,Balntas2018RelocNetCM, Zhou2019ToLO} only make use of the direction of the relative translation between the two given images.
However, by using the translation scale we could further improve the pose estimation accuracy in the triangulation step.
So, we predict the scale $\scale$ using an extra output dimension.
The loss function is adjusted accordingly: $
\poseloss = \abs{\uvec{\transl} - \uvec{\transl^*}} + \quatlossweight \abs{\quat - \quat^*} + \scalelossweight \abs{\scale - \norm{\transl^*}}$.
As we only train on indoor scenes, where the scale has a maximum value of a few meters, we use $\scalelossweight = 1$.

\PAR{Uncertainty estimation}
By using Monte Carlo dropout \cite{Gal2016UncertaintyID} we obtain uncertainty estimates which are then used to make the triangulation step described in the next section more precise.
One dropout layer is therefore placed between the two hidden fully connected layers in the last part of the network using a dropout rate of $0.1$.
At test time, we sample 100 pose estimations, use their mean as final pose estimate and the variance of the predicted camera center as uncertainty estimate $\unc_{\refI \rightarrow \query}$.

%% file: figures/network.tex
\definecolor{bbblock}{HTML}{F72C25}
\definecolor{corrblock}{HTML}{B1A306}
\definecolor{fcblock}{HTML}{F7B32B}
\definecolor{dropblock}{HTML}{020887}
\definecolor{upblock}{HTML}{A9E5BB}
\definecolor{block}{HTML}{2D1E2F}
\definecolor{poolblock}{named}{green}

\scalebox{0.6}
{\begin{tikzpicture}[
thick,
generalstyle/.style={blur shadow={shadow blur steps=5}, fill=white},
generalblock/.style={draw, rounded corners=3pt, rectangle, line width=0.5mm, generalstyle},
bbblock/.style={generalblock, draw=bbblock, minimum height=1.2cm},
corrblock/.style={generalblock, draw=corrblock, minimum height=1.2cm},
block/.style={generalblock, draw=block, minimum height=1.2cm},
fcblock/.style={generalblock, draw=fcblock, minimum height=1.2cm},
poolblock/.style={generalblock, draw=poolblock, minimum height=1.2cm},
dropblock/.style={generalblock, draw=dropblock
, minimum height=1.2cm},
upblock/.style={generalblock, draw=upblock, minimum height=1.2cm, },
concat/.style={draw, circle},
line/.style={draw, -latex,rounded corners=5pt},
legend/.style={draw, , minimum height=0.2cm, minimum width= 0.2cm}
]

\node[] (start) {};
\node[label=below:$\mathcal{I}_i$,inner sep=0pt, generalstyle] (rel_img_1) [below=1.5cm of start] {\includegraphics[height=2.0cm,width=2.0cm]{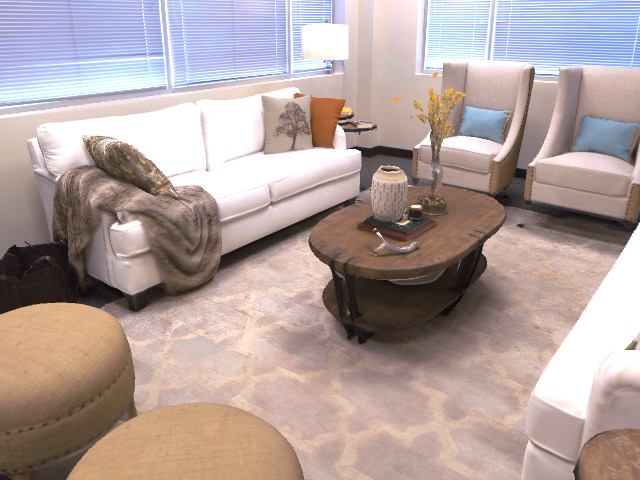}};

\node[bbblock, minimum height=1.5cm] (s1b1) [right=of rel_img_1] {$3 \times 64$};

\node[bbblock, minimum height=1.0cm] (s1b2) [right=of s1b1] {$4 \times 128$};

\node[bbblock, minimum height=0.5cm] (s1b3) [right=of s1b2] {$6 \times 256$};

\node[bbblock, minimum height=1.0cm] (s1b4) [right=of s1b3] {$1 \times 128$};

\node[bbblock, minimum height=1.5cm] (s1b5) [right=of s1b4] {$1 \times 64$};

\node[corrblock, minimum height=0.5cm] (ext_corr) [below=of s1b4] {\begin{tabular}{l} Ext. \\ corr. layer\end{tabular}};

\node[bbblock, minimum height=1.0cm] (s2b4) [below=of ext_corr] {$1 \times 128$};

\node[bbblock, minimum height=1.5cm] (s2b5) [right=of s2b4] {$1 \times 64$};

\node[bbblock, minimum height=0.5cm] (s2b3) [left=of s2b4] {$6 \times 256$};

\node[bbblock, minimum height=1.0cm] (s2b2) [left=of s2b3] {$4 \times 128$};

\node[bbblock, minimum height=1.5cm] (s2b1) [left=of s2b2] {$3 \times 64$};

\node[label=below:$\mathcal{I}_q$,inner sep=0pt] (rel_img_2) [left=of s2b1] {\includegraphics[height=2.0cm,width=2.0cm]{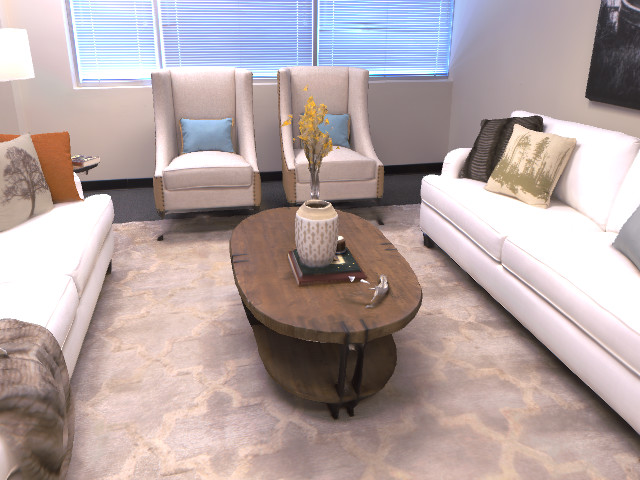}};

\node[minimum height=1.2cm] (corr_space) [right=of ext_corr] {};

\node[corrblock, minimum height=1.5cm] (guided_corr) [right=of corr_space] {\begin{tabular}{l} Guided \\ corr. layer\end{tabular}};

\node[upblock, minimum height=2cm] (guided_upscale) [right=0.5cm of guided_corr] {};

\node[upblock, minimum height=2cm] (ext_upscale) [below=of guided_upscale] {};


\node[block, minimum height=1.5cm] (regrb1) [right=1.5cm of guided_upscale] {$2 \times 64$};

\node[block, minimum height=1.0cm] (regrb2) [right=0.2cm of regrb1] {$2 \times 128$};

\node[block, minimum height=0.5cm] (regrb3) [right=0.2cm of regrb2] {$2 \times 256$};

\node[block, minimum height=0.25cm] (regrb4) [right=0.2cm of regrb3] {$2 \times 512$};
\node[poolblock, minimum height=0.5cm] (regrb5) [right=0.2cm of regrb4] {};

\node[fcblock, minimum height=1.2cm] (regrfc1) [right= of regrb4] {};

\node[dropblock, minimum height=1.2cm] (regrdrop) [right=0.2cm of regrfc1] {};

\node[fcblock, minimum height=1.2cm] (regrfc2) [right=0.2cm of regrdrop] {};

\node[fcblock, minimum height=0.5cm] (regrfc3) [right=0.2cm of regrfc2] {};

\node[] (out) [right=of regrfc3, rotate=90,anchor=north] {$(R_{i \rightarrow q}, t_{i \rightarrow q}, s_{i \rightarrow q})$};


\path [line] (rel_img_1.east) -- node[rotate=90,anchor=west]{\footnotesize $128 \times 128 \times 3$} (s1b1.west);
\path [line] (s1b1.east) --  (s1b2.west);
\path [line] (s1b2.east) --  (s1b3.west);
\path [line] (s1b3.east) --  (s1b4.west);
\path [line] (s1b4.east) --  (s1b5.west);

\path [line] (s1b1.east) to [out=40,in=140] (s1b5.west);
\path [line] (s1b2.east) to [out=40,in=140] (s1b4.west);

\path [line] (rel_img_2.east) -- node[rotate=90,anchor=east]{\footnotesize $128 \times 128 \times 3$} (s2b1.west);
\path [line] (s2b1.east) --  (s2b2.west);
\path [line] (s2b2.east) --  (s2b3.west);
\path [line] (s2b3.east) --  (s2b4.west);
\path [line] (s2b4.east) --  (s2b5.west);

\path [line] (s2b1.east) to [out=-40,in=-140] (s2b5.west);
\path [line] (s2b2.east) to [out=-40,in=-140] (s2b4.west);

\path [line] (ext_corr.east) --  node[yshift=-0.1cm, anchor=south]{\footnotesize $8 \times 8 \times 64$} (guided_corr.west);
\path [line] (guided_corr.east) --  node[rotate=90,anchor=west]{\footnotesize $32 \times 32 \times 144$} (guided_upscale.west);
\path [line] (guided_upscale.east) --  node[left=0.4cm, rotate=90,anchor=west]{\footnotesize $128 \times 128 \times 144$}  node[right=0.4cm, rotate=90,anchor=west]{\footnotesize $128 \times 128 \times 208$} (regrb1.west);

\path [line] (regrb1.east) --  (regrb2.west);
\path [line] (regrb2.east) --  (regrb3.west);
\path [line] (regrb3.east) --  (regrb4.west);
\path [line] (regrb4.east) --  (regrb5.west);

\path [line] (regrb5.east) -- node[rotate=90,anchor=east]{\footnotesize $512$}  (regrfc1.west);
\path [line] (regrfc2.east) --  (regrfc3.west);
\path [line] (regrfc1.east) --  (regrdrop.west);
\path [line] (regrdrop.east) --  (regrfc2.west);
\path [line] (regrfc3.east) -- node[rotate=90,anchor=east]{\footnotesize $8$}  (out.north);

\path [line] (s1b3.east) --  node[yshift=-0.1cm,left=0.05cm, rotate=90,anchor=east]{\footnotesize $8 \times 8 \times 1024$} +(+0.25cm,0) |-  ([yshift=.3cm]ext_corr.west);
\path [line] (s2b3.east) -- node[yshift=0.1cm,left=0.05cm, rotate=90,anchor=west]{\footnotesize $8 \times 8 \times 1024$} +(+0.25cm,0) |- ([yshift=-.3cm]ext_corr.west);

\path [line] (s1b5.east) --  node[right=0.05cm, yshift=0.6cm, rotate=90,anchor=north east]{\footnotesize $32 \times 32 \times 256$} +(+0.15cm,0) |- ([yshift=.3cm]guided_corr.west);
\path [line] (s2b5.east) --  node[right=0.05cm, yshift=-0.6cm, rotate=90,anchor=north west]{\footnotesize $32 \times 32 \times 256$} +(+0.15cm,0) |- ([yshift=-.3cm]guided_corr.west);

\path [line] (ext_upscale.east) -- +(+0.75cm,0) |- node[right=0.05cm, yshift=-2cm, rotate=90,anchor=north]{\footnotesize $128 \times 128 \times 64$} (regrb1.west);
\path [line] (ext_corr.east) -- (ext_upscale.west);

\matrix [draw,below left,xshift=-1cm,yshift=-0.3cm] at (current bounding box.north east)
{
 \node [bbblock,legend,label=right:\footnotesize ResNet bottleneck stack] {}; \\
 \node [block,legend,label=right:\footnotesize ResNet stack] {}; \\
 \node [fcblock,legend,label=right:\footnotesize Fully connected layer] {}; \\
 \node [upblock,legend,label=right:\footnotesize Upsampling layer] {}; \\
 \node [dropblock,legend,label=right:\footnotesize Dropout layer] {}; \\
 \node [corrblock,legend,label=right:\footnotesize Correlation layer] {}; \\
 \node [poolblock,legend,label=right:\footnotesize Global average pooling] {}; \\
};

\end{tikzpicture}
}

%% file: chapters/pose_estimation_pipeline/triangulation.tex
\subsection{Triangulation}

By combining two relative pose estimates $(\rot_{\refI \rightarrow \query}, \transl_{\refI \rightarrow \query})$ and $(\rot_{\refII \rightarrow \query}, \transl_{\refII \rightarrow \query})$ the absolute pose $(\rot_{\image_\query}, \transl_{\image_\query})$ of the query image $\image_q$ is estimated.
Here, we extend the absolute pose estimation step proposed by Zhou et al. \cite{Zhou2019ToLO}, which is summarized in the following paragraph.

\subsubsection{Triangulation baseline}

The camera center $\camcenter_\query$ of the query image is estimated by triangulation \cite{hartley1997triangulation} of the two rays defined by the two relative pose estimates.
For a given relative pose estimate $(\rot_{\refI \rightarrow \query}, \transl_{\refI \rightarrow \query})$ the ray is defined as $\camcenter_{\image_\refI} - \raypos_{\refI \rightarrow \query} \ray_\refI$ with the ray direction $\ray_\refI = \rot_{\image_\refI}^T  \rot_{\refI \rightarrow \query}^T \uvec{\transl}_{\refI \rightarrow \query}$ and the position $ \raypos_{\refI \rightarrow \query}$ along the ray.
The rotation $\rot_{\image_\query}$ of the query image is estimated by the average rotation between $\rot_{\refI \rightarrow \query} \rot_{\image_\refI}$ and $\rot_{\refII \rightarrow \query} \rot_{\image_\refII}$.
In this way, a hypothesis $\transform_{\hypo(\refI,\refII)}$ of the query pose is formed. 

As there might be outliers in the relative pose estimations, RANSAC is used to find a hypothesis that is consistent based on the five retrieved reference images.
Given a  hypothesis $\transform_{\hypo(\refI,\refII)}$ a reference image $\image_\refIII$ counts as an inlier, if the angle $\triangangle_\refIII$ between the direction of its ray $\ray_\refIII$ and the vector from its camera center $\camcenter_{\image_k}$ to the camera center of the hypothesis $\camcenter_{\hypo(\refI,\refII)}$ is smaller than a threshold $\triangangle_{max}$. 
Finally, the hypothesis with the most inliers is used.

\subsubsection{Our adaptation}

We found that using only the angle $\triangangle_\refIII$ to determine
inliers can often lead to false positives. To prevent such cases, we
 propose to additionally use the translation scale estimation
$\scale_{\refIII \rightarrow \query}$ for determining inliers.  Thus,
in addition to the constraint of $\triangangle_\refIII < \triangangle_{max}$,
a reference image $\image_\refIII$ has to also fulfill the criteria
$\scale_{min} < \scale_\hypo / \scale_{\refIII \rightarrow \query} <
\scale_{max}$ to count as an inlier for a hypothesis
$\transform_{\hypo(\refI,\refII)}$.  Here, $\scale_{\refIII
  \rightarrow \query}$ is the predicted scale between the query image
$\image_\query$ and the reference image $\image_k$ and $\scale_\hypo$
is the actual scale of the relative translation between its camera
center $\camcenter_{\image_\refIII}$ to the camera center of the
hypothesis $\camcenter_{\hypo(\refI,\refII)}$.  For all our
experiments we used $\triangangle_{max} = 15$\degree, $\scale_{min} =
0.5$ and $\scale_{max} = 2.0$.

To further improve the absolute pose estimation, we additionally take
the uncertainty estimate $\unc_{\refIII \rightarrow \query}$ of the
network into account: We use uncertainty information to
disambiguate cases where multiple hypotheses have the same number
of inliers.  If that is the case, we use the hypothesis
$\transform_{\hypo(\refI,\refII)}$ with the smallest mean uncertainty
$0.5\parenth{\unc_{\refI \rightarrow \query} + \unc_{\refII
    \rightarrow \query}}$ of both corresponding relative pose
estimations.

%% file: chapters/experiments.tex
\section{Experiments}

\subsection{General procedure}

\captionsetup{position=t}

\begin{table*}[ht!]
\vspace{2mm}
   \scriptsize
	\caption[Comparison of a big variety of approaches on the 7-Scenes dataset]{Comparison of a big variety of approaches on the 7-Scenes dataset: The models are compared based on their median translation and rotation errors averaged across all scenes.}
	\label{table:big_table}
\subfloat[Methods trained on 7-Scenes or requiring 3D models for each scene]{
 \begin{minipage}{.39\linewidth}
	\centering
	\begin{tabular}{l   l l  }\toprule
    & Scene-specific & Errors  \\ 
    \midrule     
    \multicolumn{3}{c}{3D structure-based}\\    
    \midrule    
    Active Search \cite{7572201}  & Yes & 0.05m / 2.46\degree \\
    DSAC++ \cite{brachmann2018learning} & Yes & 0.04m / 1.10\degree \\   
    \midrule 
    \multicolumn{3}{c}{Absolute pose estimation}\\     
    \midrule    
    PoseNet \cite{DBLP:journals/corr/KendallGC15} & Yes & 0.44m / 10.44\degree \\
    Geo. PN \cite{Kendall2017GeometricLF}  & Yes & 0.23m / 8.12\degree \\
    MapNet \cite{Brahmbhatt2017GeometryAwareLO} & Yes & 0.21m / 7.78\degree \\
    \midrule
    \multicolumn{3}{c}{Relative pose estimation}\\  
    \midrule   
    Relative PN \cite{Laskar2017CameraRB} & No & 0.21m / 9.28\degree \\
    RelocNet \cite{Balntas2018RelocNetCM} & No & 0.21m / 6.73\degree \\
    EssNet \cite{Zhou2019ToLO} & No & 0.22m / 8.03\degree 
     \\
    NC-EssNet \cite{Zhou2019ToLO} & No & 0.21m / 7.50\degree \\
    AnchorNet \cite{Saha2018ImprovedVR} & Yes & 0.09m / 6.74\degree  \\ 
    CamNet \cite{ding2019camnet} & No & 0.04m / 1.69\degree  \\
    \midrule   
 	\label{table:compare_spec}
    \end{tabular}
	
	\end{minipage}}
	\qquad
\subfloat[Methods trained on unrelated data or with no training at all]{
	 \begin{minipage}{.59\linewidth}
	\centering
	  \begin{tabular}{l l l l  l l  }\toprule
    \multicolumn{3}{l}{} & Input size & Traindata & Errors  \\ 
    \midrule    
    \multicolumn{6}{c}{Image retrieval} \\    
    \midrule  
    \multicolumn{3}{l}{DenseVLAD \cite{torii201524}} & $640 \times 480$ & - &  0.26m / 13.11\degree \\
    \midrule
    \multicolumn{6}{c}{Relative pose estimation}\\  
    \midrule    
    \multicolumn{3}{l}{Sift+5pt \cite{Zhou2019ToLO}}& $640 \times 480$ & - & 0.08m / 1.99\degree  \\
    \multicolumn{3}{l}{Sift+5pt} & $149 \times 110$ & - & 0.13m / 2.86\degree \\
    \multicolumn{3}{l}{Relative PN \cite{Laskar2017CameraRB}} & $224 \times 224$ & University & 0.36m / 18.37\degree \\
    \multicolumn{3}{l}{RelocNet \cite{Balntas2018RelocNetCM}} & $224 \times 224$ & ScanNet & 0.29m / 11.29\degree \\
    \midrule
    \multicolumn{6}{c}{Ours}\\
    \midrule    
    EssNet \cite{Zhou2019ToLO} & & & $224 \times 224$ &  SUNCG &  0.25m / 9.76\degree \\
     ExReNet & Scale: \xmark & Unc: \xmark & $128 \times 128$ &  ScanNet & 0.12m / 3.30\degree \\ 
    ExReNet & Scale: \xmark & Unc: \xmark & $128 \times 128$  & SUNCG & 0.11m / 2.97\degree \\ 
    ExReNet & Scale: \cmark & Unc: \xmark & $128 \times 128$ & SUNCG & 0.10m / 2.98\degree \\ 
    ExReNet & Scale: \cmark & Unc: \cmark & $128 \times 128$ &  SUNCG& 0.09m / 2.72\degree \\ 
   
    \bottomrule
	\label{table:compare_agn}
    \end{tabular}
	\end{minipage}
	}
	
\end{table*}

\captionsetup{position=b}

If not stated otherwise, all our networks are trained only on synthetic data generated using BlenderProc\cite{denninger2019blenderproc}.
As 3D models, we select the 1500 most furnished houses from the SUNCG dataset \cite{song2017semantic}, in which we randomly sample valid camera poses.
We use two different synthetic training sets, one for densely and one for sparsely covered scenes.
While we sample for the dense training set camera poses along all six DoF, for the sparse training set we mostly only vary along three degrees of freedom (x, y and yaw angle), which is done to simulate a moving robot.
 
The network itself is trained on randomly sampled image pairs from the respective training set, whereby we require that the overlap $\intersection_{\refI,\refII}$ is bigger than a configured threshold $\intersecmin$, the distance between the camera centers $\norm{\camcenter_{\image_\refI} - \camcenter_{\image_\refII}}$ is smaller than $\distancemax$, and that the rotation difference $\gamma$ is smaller than a threshold $\anglemax$.
Here, the overlap factor $\intersection_{\refI,\refII}$ of two given images $\image_\refI$ and $\image_\refII$ is defined as the ratio of scene coordinates that are located at most $\scenecoordneighbor=0.2m$ away from a scene coordinate from the other image.
For the dense training set we use $\intersecmin = 30\%$, $\distancemax = 0.6m$ and $\anglemax = 30$\degree.
For the sparse training set we use $\intersecmin = 10\%$ as the only condition.
During training, all images are randomly augmented by applying random brightness, contrast and saturation. 
All models are trained using the Adam optimizer using a learning rate of $5\mathrm{e}{-5}$ for 500k iterations.

\input{chapters/experiments/dense}

\input{chapters/experiments/sparse}

%% file: chapters/experiments/dense.tex
\subsection{Dense reference image coverage}

In this section, we evaluate ExReNet on scenes that are densely covered by reference images in the form of the commonly used 7-Scenes dataset.
In the case of scene-agnostic approaches, like ours, the training images of 7-Scenes are used as reference images during inference.

\subsubsection{Generalization to unseen scenes}

Table \ref{table:compare_spec} and \ref{table:compare_agn} list the pose estimation accuracy of a variety of approaches on 7-Scenes.
It is apparent that 3D structure-based methods still outperform all other approaches.
However, as stated earlier, they need either a detailed 3D model or require extensive scene-specific training.
Apart from that, our scene-agnostic approach outperforms all methods trained on unrelated data and also most of the approaches that were specifically trained on 7-Scenes.
Especially, the rotational error of our network is much lower compared to networks trained on 7-Scenes, and it is more in the range of classical methods.
Using scale and uncertainty information further improves especially the translation error, which is reasonable as both extensions are used for the estimation of the camera center of the query image.
In this way, we outperform handcrafted SIFT features together with the 5-point algorithm on an equivalent resolution. 
This also stays true, if the number of reference images is artificially reduced (see Fig.~\ref{fig:ref_images_7scenes}).
To better understand the influence of synthetic data on these results, we also trained our model on ScanNet which surprisingly results in a slightly worse performance compared to using synthetic data.
We relate this phenomenon to the fact that the images in ScanNet\cite{dai2017scannet} are not sampled uniformly as in our synthetic dataset, but are recorded on trajectories.

\subsubsection{Overfitting effect}
\label{sec:exp_7scenes_visualization}

\begin{figure*}[h]

\subfloat[EssNet trained on 7-Scenes]{
  \begin{minipage}[b]{0.3\textwidth}
    \includegraphics[width=0.8\textwidth]{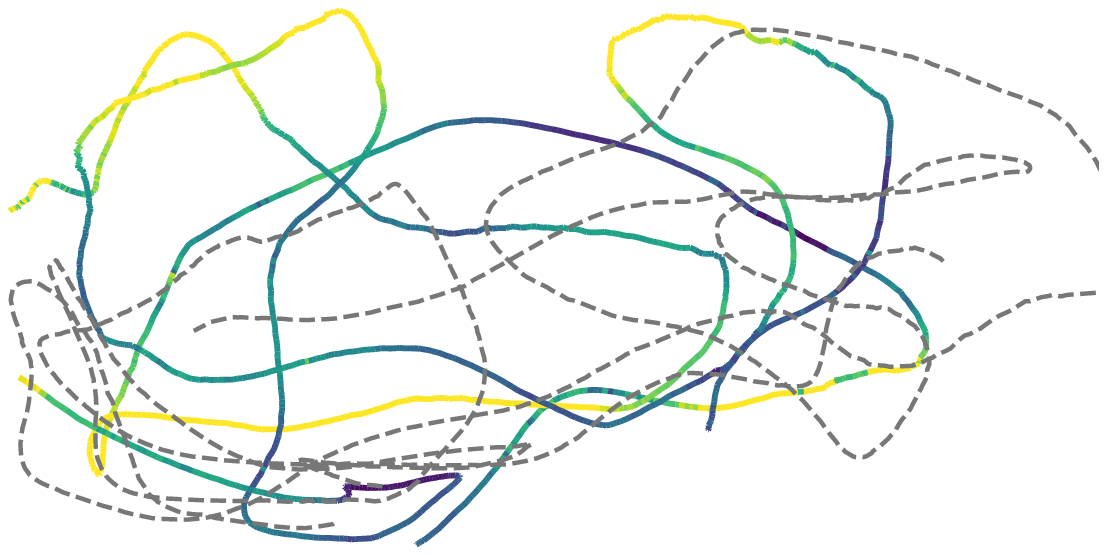}
  \end{minipage}
}
\subfloat[ExReNet trained on 7-Scenes]{
  \begin{minipage}[b]{0.3\textwidth}
    \includegraphics[width=0.8\textwidth]{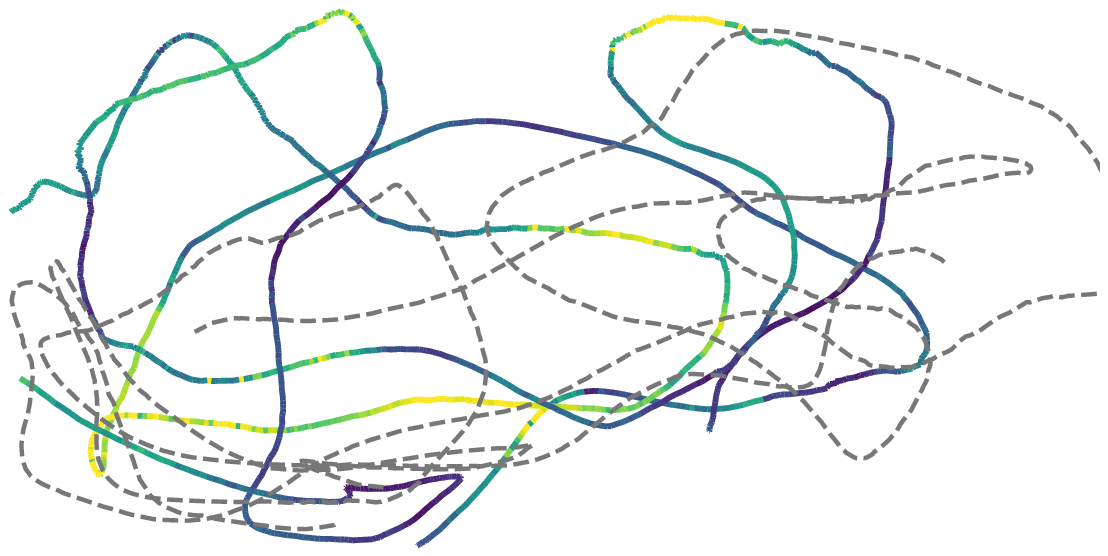}
  \end{minipage}
}
\subfloat[ExReNet trained only on synthetic data]{
  \begin{minipage}[b]{0.3\textwidth}
   \includegraphics[width=0.8\textwidth]{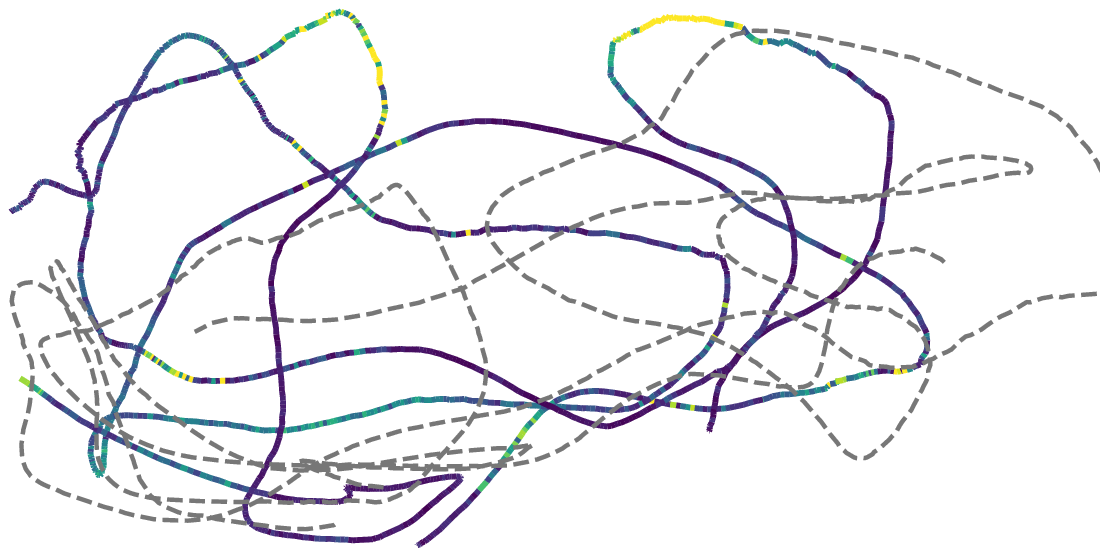}
  \end{minipage}
}
\subfloat{
  \begin{minipage}[t]{0.10\textwidth}
   \includegraphics[height=2cm]{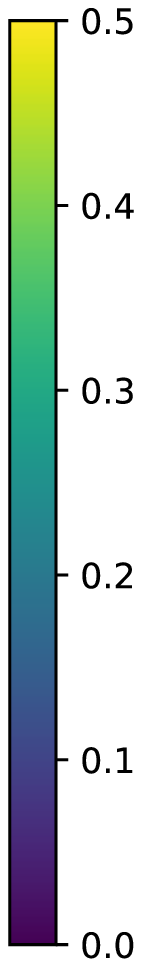}
  \end{minipage}
}

\caption{Visualization of the overfitting effect of relative pose
  estimation networks. Here, we show the deviations between predicted
  and ground truth camera center on the fire scene from 7-Scenes. The
  gray dashed line represents the 7-Scenes reference images (training data for a and b), while the
  query images are visualized with a color depending on the prediction error. The mapping between deviation in meters and color
  is shown in the bar on the right.}
\label{fig:error_traj}
\end{figure*}

By further investigating why our network trained on synthetic data outperforms many methods trained on 7-Scenes itself, we discovered similar results as had been shown for absolute pose estimation methods  \cite{sattler2019understanding}:
Relative pose estimation methods also tend to overfit on training trajectories.
We visualize this phenomenon in Fig. \ref{fig:error_traj}.
Here, we compare the errors of the predicted camera centers between an EssNet trained on 7-Scenes, an ExReNet also trained on 7-Scenes and one only trained on synthetic data.
It is clearly visible that the error of the predictions made by EssNet increases if the test images are further away from the training trajectories.
When training ExReNet on 7-Scenes this phenomenon is weaker but still present.
However, it cannot be observed in the figure based on ExReNet trained on synthetic data.
From that we conclude that also relative pose estimation methods can overfit on the training trajectories, and the training on unrelated synthetic data that incorporates a big variety of camera poses helps the generalization.

\subsubsection{Ablation studies}
\label{sec:exp_7scenes_regressors}

\begin{table}[h!]
   \scriptsize
	\caption{Results of training ExReNetClassic with different feature matching methods and regression parts.}
	\centering
	\begin{tabular}{l  l }\toprule
    &  Avg. of median errors \\ 
    \midrule
    ExReNet & 0.11m / 2.97\degree \\ 
    ExReNetClassic (one corr. layer on $8 \times 8$) & 0.15m / 4.19\degree \\ 
    \midrule
    ExReNetClassic with no ResNet blocks in regressor &  0.39m / 9.01\degree \\ 
    ExReNetClassic with EssNet-like regression part & 0.30m / 7.22\degree \\
    \midrule
    ExReNetClassic with concatenation channel-wise & 0.25m / 5.54\degree  \\ 
    ExReNetClassic with concatenation after pooling & 0.36m / 7.89\degree  \\  
    \bottomrule
    \end{tabular}
	\label{table:regressor_comparison}
\end{table}

\begin{table}[t!]
\scriptsize
	\caption{Comparison of multiple approaches on the sparse covered scenes: This error considers query images where a prediction was possible. The ratio of images for which a prediction was possible is given in brackets behind each error.}
	\centering
	\begin{tabular}{ l @{\hspace{0.2cm}} l @{} l @{\hspace{0.2cm}} l @{} l}\toprule
     & \multicolumn{2}{c}{replica-based dataset} & \multicolumn{2}{c}{custom real-world data} \\ 
    \midrule
    DenseVLAD & 2.85m / 88.49\degree & (100 \%) & 1.80m / 19.26\degree & (100 \%) \\ 
    Sift+5pt + IR fallback & 2.16m / 58.11\degree & (100 \%) & 1.62m / 20.09 \degree &  (100 \%) \\ 
    EssNet & 1.50m / 160.98\degree & (100 \%) & 1.10m / 125.09\degree & (100 \%)\\
    \midrule
    Sift+5pt (Lowe's ratio: 0.8) & 0.50m / 18.68\degree & \textcolor{red}{(36 \%)}  & 0.65m / 10.66\degree & \textcolor{red}{(26\%)} \\ 
    Sift+5pt (Lowe's ratio: 0.9) & 1.66m / 104.91\degree & \textcolor{red}{(75 \%)} & 1.05m / 108.3\degree & \textcolor{red}{(79\%)} \\ 
    \midrule
    ExReNet Scale: \xmark, Unc: \xmark  & 0.92m / 29.74\degree  &(100 \%)  & 0.64m / 10.00\degree & (100 \%) \\    
    ExReNet Scale: \cmark, Unc: \xmark   & 0.77m / 17.22\degree  & (100 \%) & 0.61m / 8.37\degree & (100 \%)\\ 
    ExReNet Scale: \cmark, Unc: \cmark   & 0.64m / 9.39\degree  &(100 \%) & 0.61m / 7.74\degree & (100 \%) \\ 
    \bottomrule
    \end{tabular}
	\label{table:replica}
\end{table}

The fact that EssNet, trained on our synthetic training data, still results in bad generalization (see table \ref{table:compare_agn}) hints that our proposed architectural changes also have beneficial effects:
As can be seen in table \ref{table:regressor_comparison}, when using only one correlation layer on $8 \times 8$ feature maps, the accuracy decreases, however, it is still more accurate than most methods listed in table \ref{table:big_table}.
If the regressor is scaled down further, the model performs considerably worse.
So, it seems that using a larger regressor, that can fully learn the complex mapping from the feature matches to the resulting relative pose, has a stronger ability to generalize.
Furthermore, the ablation study confirms that the performance decreases considerably when replacing the correlation layer with concatenation.

%% file: chapters/experiments/sparse.tex
\subsection{Sparse map}

\begin{figure}[h]

\subfloat[7-Scenes]{
   \begin{minipage}[t]{0.5\linewidth}
    \centering
   \includegraphics[width=1\linewidth]{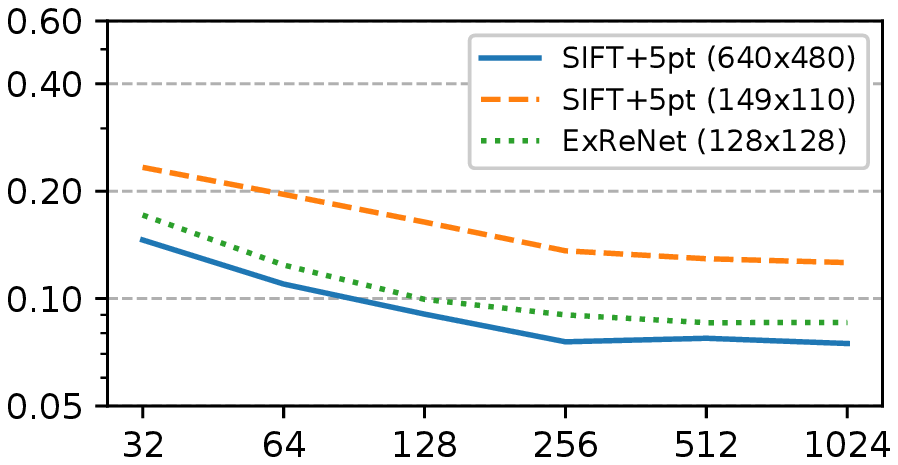}
  \end{minipage}
  \label{fig:ref_images_7scenes}
}
\subfloat[Our recorded dataset]{
   \begin{minipage}[t]{0.5\linewidth}
    \centering
   \includegraphics[width=1\linewidth]{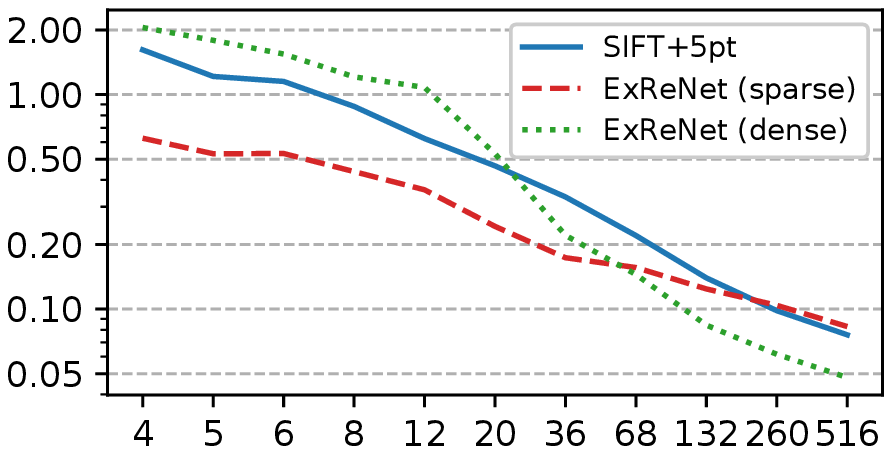}
  \end{minipage}
  \label{fig:ref_images_smile}
}

  \caption{Visualizes how the translational error changes depending on the number of reference images per scene. On 7-Scenes the reference images per scene are randomly reduced, while on our recorded dataset random reference images are added additional to the four ones used in table \ref{table:replica}. ExReNets are evaluated using scale and uncertainty information.}
\label{fig:ref_images}
\end{figure}

In contrast to 7-scenes, where each scene is densely covered with reference images, we evaluate here the case where only four reference images per indoor scene are given. 
Each of them is placed in one corner of the room, oriented towards the center.
We evaluate this scenario on two datasets:
One is based on replica \cite{replica19arxiv}, where we manually placed the four reference cameras in every room and sample 500 query poses the same way as for the sparse training set.
Using a second dataset, we show that this also works on images recorded from a sensor of our robotic system.
Therefore, an Intel Realsense d435 sensor is used to record multiple trajectories in our kitchen setup.
The ground truth poses are determined by applying ORBSLAM2 \cite{murORB2} to the recorded trajectories.
For comparison, ORBSLAM2, when evaluated on 7-Scenes, achieves an error of 0.06m / 2.55\degree, which makes it comparable to other 3D structure-based methods.
Again, we extract four reference images this time from the recorded trajectories, one per corner facing towards the center.
The results in table \ref{table:replica} show that image retrieval performs worst, which is reasonable due to the low number of reference images.
Using Sift+5pt in the same way as with the 7-Scenes dataset leads to a big number of cases where no prediction is possible at all.
Especially if the baseline between a given query reference pair is very large, the number of successful feature matches is too low to estimate the relative pose via the 5-point algorithm.
By increasing Lowe's ratio, we can increase the number of feature matches we mark as correct, but at the same time also the accuracy of the predicted poses decreases due to the lower quality of feature matches.
Compared to all tested classical and learning-based methods, our approach performs best.
This shows that in the case of only a few reference images and thus large baselines between query and reference images, learning-based approaches can outperform classical non-learning based approaches in scene-agnostic pose estimation.
As can be seen in Fig.~\ref{fig:ref_images_smile}, this stays true if further additional reference images are given.
Starting from 64 additional reference images the ExReNet trained on dense training data used in the last section can be used to achieve the best accuracy.

%% file: chapters/conclusions.tex
\section{Conclusions}
\label{sec:conclusions}

In this paper we showed that relative pose estimation can be used to
localize in new scenes without retraining or detailed 3D models.  By
extending the regression part and adding a guided correlation layer,
we create a novel relative pose estimation architecture called
ExReNet, which is able to generalize to new scenes.  Additionally, we
demonstrated how translation-scale and uncertainty information can be
further used to improve the results.  On sparse and dense covered
datasets such as 7-Scenes our approach beats SIFT+5pt and outperforms
all methods trained on unrelated data and most approaches directly
trained on the evaluation scene, while our network is only trained on
unrelated synthetic data.